\documentclass[journal]{IEEEtran}

\usepackage{cite}
\usepackage{amsmath,amssymb,amsfonts}
\usepackage{algorithm}
\usepackage{algorithmic}
\usepackage{graphicx}
\usepackage{epstopdf}
\usepackage{textcomp}
\usepackage{array}
\usepackage{hyperref}
\usepackage{multirow}
\usepackage{url}
\usepackage{upgreek}
\usepackage{subfigure}
\usepackage{setspace}
\usepackage{siunitx}
\usepackage{comment}
\usepackage{etoolbox}
\usepackage{units}
\usepackage{bm}
\usepackage{etoolbox}
\usepackage{balance}
\usepackage{color}
\makeatletter
\patchcmd{\@makecaption}
  {\scshape}
  {}
  {}
  {}
\makeatother
\ifCLASSINFOpdf
\else
\fi
\begin{document}

\title{HVAQ: A High-Resolution Vision-Based Air Quality Dataset}

\author{	Zuohui~Chen$^{\ast}$,
            Tony~Zhang$^{\ast}$,
            Zhuangzhi~Chen,
            Yun~Xiang$^{\dag}$,
            Qi~Xuan,
            Robert~P.~Dick% <-this % stops a space
\thanks{ % College of Information Engineering, Zhejiang University of Technology, Hangzhou 310023, China
Z. Chen, Z. Chen, Y. Xiang ($^{\dag}$corresponding author, email: xiangyun@zjut.edu.cn), and Q. Xuan are with the Institute of Cyberspace Security, Zhejiang University of Technology, Hangzhou 310023, China.
R. Dick and T. Zhang are with the EECS Department, University of Michigan, Ann Arbor, U.S.A.
}% <-this % stops a space
\thanks{Authors with $^{\ast}$ contributed equally to this work.
}% <-this % stops a space
\thanks{}}



\maketitle

\begin{abstract}
Air pollutants, such as particulate matter, negatively impact human health. Most existing pollution monitoring techniques use stationary sensors, which are typically sparsely deployed. However, real-world pollution distributions vary rapidly with position and the visual effects of air pollution can be used to estimate concentration, potentially at high spatial resolution. Accurate pollution monitoring requires either densely deployed conventional point sensors, at-a-distance vision-based pollution monitoring, or a combination of both.

The main contribution of this paper is that to the best of our knowledge, it is the first publicly available, high temporal and spatial resolution air quality dataset containing simultaneous point sensor measurements and corresponding images. The dataset enables, for the first time, high spatial resolution evaluation of image-based air pollution estimation algorithms. It contains PM2.5, PM10, temperature, and humidity data. We evaluate several state-of-art vision-based PM concentration estimation algorithms on our dataset and quantify the increase in accuracy resulting from higher point sensor density and the use of images. It is our intent and belief that this dataset can enable advances by other research teams working on air quality estimation. Our dataset is available at \url{https://github.com/implicitDeclaration/HVAQ-dataset/tree/master.}

%(1) it presents a new, high spatial- and temporal-resolution dataset that contains measurements from dense, stationary particle counters and images covering those particle counters; (2) we evaluate and compare state-of-art pollutant estimation algorithms using our dataset; and (3) we show how the dataset can be used to develop a new vision-based estimation algorithm with 16.9\% higher accuracy than existing techniques. It is our intent and belief that this dataset can enable advances by other research teams working on air quality estimation.

\end{abstract}

% Note that keywords are not normally used for peer review papers.

\IEEEpeerreviewmaketitle

\section{Introduction}
\label{sec:introduction}

\IEEEPARstart{A}{ir} pollution is a serious threat to human health, and is closely related to disease and death rates~\cite{naddafi2012health,nourmoradi2016air,poduri2010visibility,chen2008air,kelly2015air}. According to the World Health Organization (WHO), air pollution causes seven million premature deaths per year, largely as a result of increased mortality from stroke, heart disease, and lung cancer. The most common harmful air pollutants are particulate matter (PM), sulfur dioxide, and nitrogen dioxide. This work focuses on PM, which can increase the rate of cardiovascular and respiratory diseases~\cite{feng2016health, anderson2012clearing}.

\begin{figure} [!t]
\centering
\includegraphics[scale=0.5]{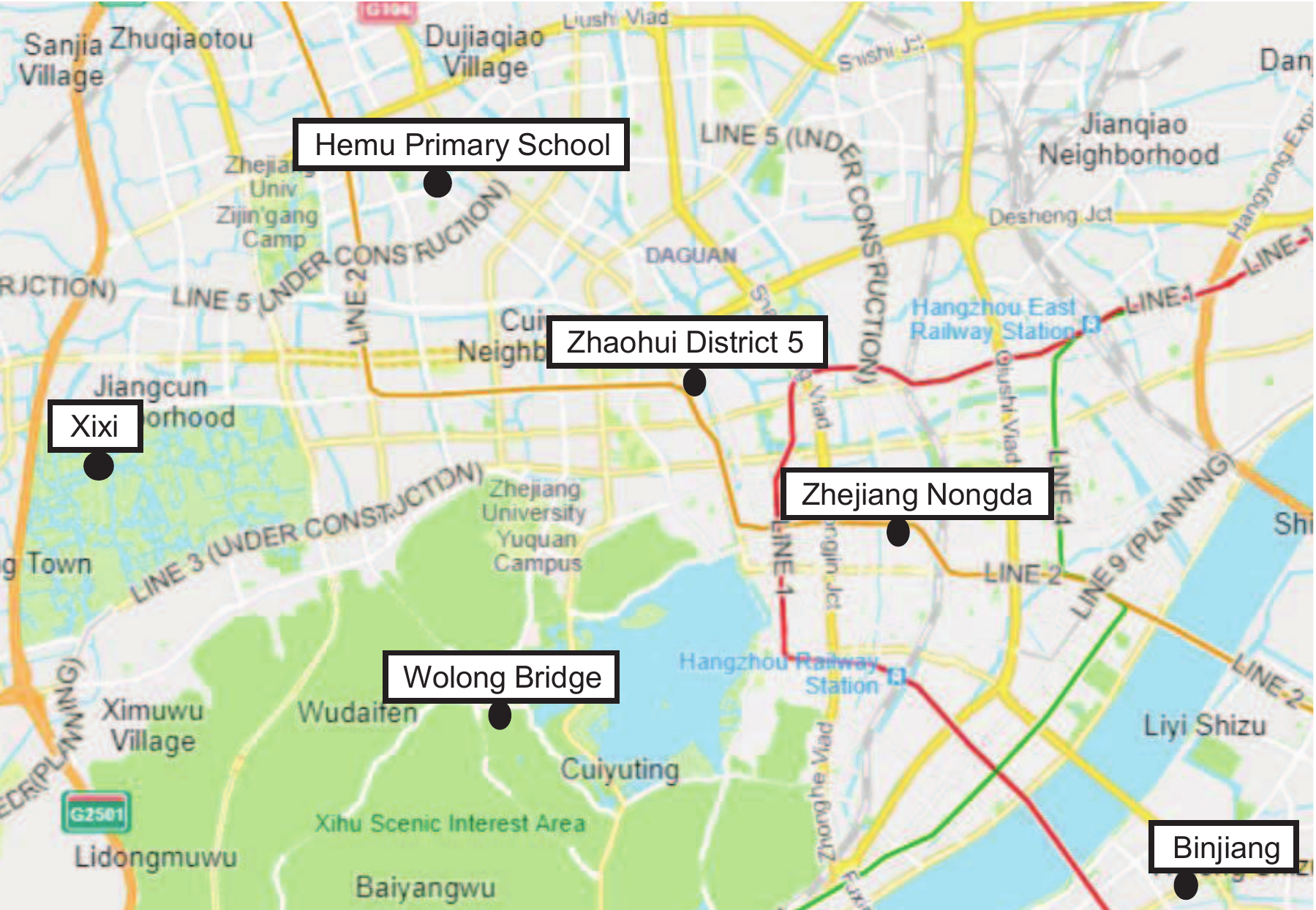}
\caption{The distribution of air monitoring stations in Hangzhou. On average, there is one monitoring site for every 35 square kilometers.}
\label{Fig:HZ_AMS}
\end{figure}

Air monitoring stations are mainly used to obtain air pollution data~\cite{rohde2015air}. However, air pollution concentration can vary within a relatively short distance: the low sensor density (see \autoref{Fig:HZ_AMS}) can lead to inaccurate estimation of the high-resolution pollution field. Low measurement spatial density makes it especially difficult to estimate human exposure to air pollution.

The PM concentrations are correlated with source distributions. For example, PM has heterogeneous sources~\cite{zhang2017review}, e.g., automobiles, manufacturing, and building construction etc. In addition, numerous factors, including wind, humidity, and geography~\cite{lin2014spatio,zhao2019spatiotemporal}, are related to PM distributions. Increasing sensor density or adding image sensors supporting high spatial resolution captures can increase the field estimation accuracy and resolution of pollutant concentrations.

%\textcolor{red}{Since real-world pollution distributions vary rapidly in space, accurate pollution monitoring requires either densely deployed conventional point sensors, at-a-distance vision-based pollution monitoring, or a combination of both. Most existing pollution monitoring techniques use stationary sensors, which are typically too sparsely deployed for accurate field estimation. In contrast to stationary particle counters, consumer-grade cameras are inexpensive and gather pollution data over a wide field of view (see Figure ~\ref{Fig:picture view}).}

\begin{figure} [!t]
\centering
 \subfigure[]
 {
  \begin{minipage}{0.24\textwidth}
   \centering
   \includegraphics[scale=0.18]{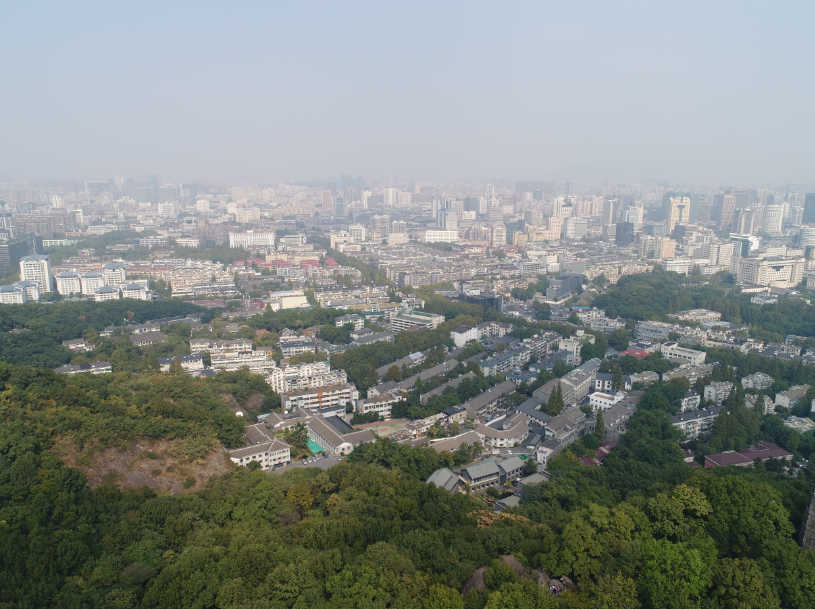}
  \end{minipage}%
  }%
 \subfigure[]
  {
  \begin{minipage}{0.24\textwidth}
   \centering
   \includegraphics[scale=0.18]{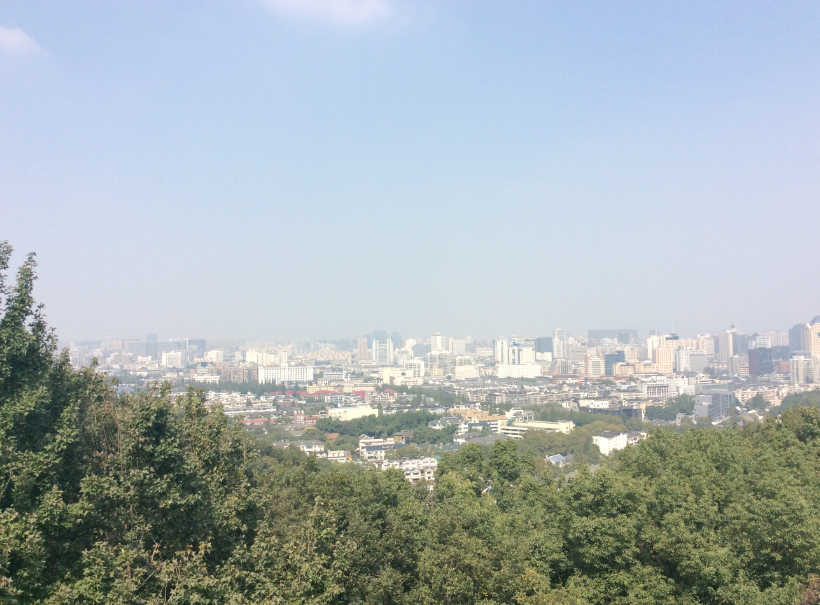}
  \end{minipage}
  }
\caption{(a) High altitude quadcopter view and (b) low altitude camera view.}
\label{Fig:picture view}
\end{figure}

In addition to air quality sensors, image sensors can also be used to estimate PM concentrations, potentially at higher spatial resolutions~\cite{zhang2016estimating,zhang2019estimation}. Image data can  be derived from several sources such as social media and digital cameras~\cite{li2015using,liu2016particle}. Moreover, consumer-grade cameras are inexpensive and can gather pollution data over a wide field of view (see Figure ~\ref{Fig:picture view}). PM can be estimated by analyzing the visual haze effect caused by scattering and absorption~\cite{zhang2019estimation}. On days with high pollution,visibility is low because light is scattered away from the camera by PM.

The ground truth pollution data are typically obtained from the nearest of several sparsely deployed monitoring stations. Lack of high-resolution ground truth pollution measurements makes it difficult to evaluate vision-based pollution estimation techniques, as pollution can vary rapidly with position~\cite{liu2016particle}. Moreover, vision-based pollution estimation techniques generally assume homogeneous distributions of particles and gases within images, implying constant light attenuation~\cite{liu2016particle}. In reality, pollution concentration changes rapidly in space. Therefore, accurate evaluation of vision-based estimation algorithms requires high-resolution datasets~\cite{apte2017high}.

This paper presents a dataset containing high spatial (one sensor every 2.5 square kilometers) and temporal (one second interval) resolution particle counter based pollution measurements with corresponding images, in addition to auxiliary information including GPS locations, humidity, and temperature. These properties are significant: this is the first publicly available dataset capable of being used to train and evaluate vision-based pollution estimation and forecasting techniques at high spatial resolutions. To the best of our knowledge, there have been no publicly available datasets enabling evaluation in this context. Certainly, There are datasets containing high-resolution~\cite{apte2017high,guan2020fine} and wide coverage~\cite{wei2021chinahighpm10,khan2019development} air pollution data. However, none of them contain corresponding synchronized images. Based on the dataset, which is the primary contribution of this paper, we also make several observations, e.g., the rate of spatial variation in pollution concentration and the evaluation of several existing vision-based PM concentration predication algorithms. To evaluate the improvement brought by increasing sensor spatial resolution and using images, we use heterogeneous information for concentration estimation, i.e., images and particle counter based PM concentrations, which were not considered in the previous vision-based algorithms.

The rest of the paper is organized as follows. \autoref{sec:related work} summarizes related work. Sections~\ref{sec:procedure} and \ref{sec:dataset} describe our data collection and analysis process. \autoref{sec:exp} presents the experimental results. \autoref{sec:conclusion} concludes the paper.

\section{Related Work}
\label{sec:related work}
The related work can be generalized into three categories: environment monitoring methods, spectrum RGB cameras, and vision-based techniques, respectively.

\begin{table}[!t]\footnotesize
    \centering
	\caption{Datasets comparison.}
	\label{Tab:datasets}
    \centering
	\begin{tabular}{|c|c|c|c|c|}%{c|>{\centering}p{0.7cm}>{\centering}p{0.7cm}>{\centering}p{0.7cm}>{\centering}p{0.7cm}>{\centering}p{0.7cm}}
		\hline
		     & Spatial & Temporal & \multirow{2}{*}{Scale} & \multirow{2}{*}{Images}  \tabularnewline
             & resolution & resolution & ~ & ~ \tabularnewline
        \hline
        Apte et al.  & High & Second & \SI{30}{\square \kilo\metre} & No \tabularnewline
        \hline
        ChinaHighPM10  & Low & Hour & 9.6M\SI{}{\square \kilo\metre} & No \tabularnewline
        \hline
        ImgSensingNet  & High & Hour & \SI{7}{\square \kilo\metre} & Yes \tabularnewline
        \hline
        Air monitoring   & \multirow{2}{*}{Low} & \multirow{2}{*}{Hour} &  \multirow{2}{*}{\SI{16853}{\square \kilo\metre}} & \multirow{2}{*}{No} \tabularnewline
        stations (Hangzhou) & ~ & ~ & ~ & ~ \tabularnewline
        \hline
        HVAQ  & High & Second & \SI{1}{\square \kilo\metre} & Yes \tabularnewline
        \hline
	\end{tabular}
\end{table}

\subsection{Environment Monitoring Methods}

Most existing air quality monitoring systems~\cite{wen2019novel} have low temporal and spatial resolutions. For example, Janssens-Maenhout et al.~\cite{janssens2012edgar} provide a harmonized gridded air pollution emission dataset. This dataset includes multiple pollutants on a global scale with $0.1^\circ \times 0.1^\circ$ spatial resolution (latitude and longitude). De et al.~\cite{de2008field} collect an air quality dataset containing 9,358 instances of hourly averaged responses from metal oxide chemical sensors. Devices are deployed in a highly polluted area in Italy at high spatial resolution.  Li et al.~\cite{li2012sensing} provide a dataset of mobile air quality measurements in Zurich. They use sensor boxes installed on mobile trams. Static installations are also deployed close to high-quality reference stations for calibration. Their sensors move around the city, recording every \SI{5}{\second}. Apte et al.~\cite{apte2017high} use two Google street view vehicles equipped with data acquisition platforms to collect the air pollution data of a \SI{30}{\square \kilo\metre} city area. Their data contains nitrogen oxides and black carbon. Unlike our dataset, it does not contain PM concentrations, images, and weather conditions. Wei et al.~\cite{wei2021chinahighpm10} describe the ChinaHighPM10 dataset, which integrates multiple data sources and contains hourly PM10 data in China with \SI{1}{\kilo\metre} resolution. Generally, these datasets typically do not include images for the evaluation of vision-based pollution estimation.

Yang et al.~\cite{yang2019imgsensingnet} describe ImgSensingNet, an air quality monitoring system consisting of unmanned-aerial-vehicles (UAV) and ground sensors. Their system uses UAVs for vision-based air quality index (AQI) monitoring and activates the ground sensing network based on vision-based AQI inference. Their dataset differs from ours as follows. (1) It is not public nor can be expected to be public in the near future. (2) Their data have time gaps, i.e., there is ground sensor data only when the UAV is uncertain about the AQI in some areas, making the intervals without ground sensor data useless for validating vision-based algorithms. In contrast, our data acquisition system operated continuously. (3) Their dataset contains an irreversible AQI index inferred from PM data, while our dataset provides raw PM data, meteorology information, and environmental information. (4) Their point sensor data acquisition period is one hour (at best, depending on whether the point sensors are activated) while ours is one second. \autoref{Tab:datasets} compares different datasets. Our HVAQ is unique in providing high spatial and temporal resolution pollution measurements with corresponding images. Moreover, it has been publicly released.

\subsection{Spectrum-Based RGB Cameras}

Image data are typically gathered using spectral measurements, imaging systems, and multi-spectral imaging~\cite{herzog2003multispectral}. Spectral measurements acquire accurate color data without structural information from the scene. Imaging systems such as scanners or digital RGB cameras capture detailed structural information~\cite{martinez2003characterization} with a limited number of color channels. Mismatches between the camera's spectral sensitivities and desired color matching functions can reduce the camera's color quality.

In contrast, multi-spectral imaging can produce accurate, detailed representations of the scene for both color and structure~\cite{nahavandi2016metric}.  Nonetheless, multi-spectral imaging systems are expensive, while spectral measurements provide very limited spatial resolution.

\subsection{Image-based Techniques}

Vision-based pollution estimation is being studied due to its spatial range and low cost. Li et al.~\cite{li2015using} analyze photos acquired from social media and establish the correlation between visual haze and PM2.5 concentration. Liu et al.~\cite{liu2016particle} estimate PM2.5 using support vector regression on six features extracted from images. Zhang et al.~\cite{zhang2016estimating} describe an air pollution estimation algorithm that uses outdoor photos. They use a convolutional neural network to predict the pollution level. Rijal et al.~\cite{rijal2018ensemble} integrate the prediction of multiple neural networks to calculate the PM2.5 of images. The ensemble of neural networks provides more accurate estimation than a single network. Gu et al.~\cite{gu2018highly} designed a vision-based estimation of PM2.5 concentration. It requires low-pollution reference images for each area of interest. Zhang et al.~\cite{zhang2019estimation} estimate multiple pollution concentrations from images using scattering and absorption models.

The main problem is that it is unclear how accurate these methods are because, until now, ground-truth data are of very low spatial resolution. Therefore, our dataset makes it possible to evaluate them in a high-resolution scenario.

\begin{table}[!t]\footnotesize
    \centering
	\caption{GPS Locations of Our Sensors and Picture.}
	\label{Tab:GPS}
    \centering
	\begin{tabular}{|c|c|c|c|}%{c|>{\centering}p{0.7cm}>{\centering}p{0.7cm}>{\centering}p{0.7cm}>{\centering}p{0.7cm}>{\centering}p{0.7cm}}
		\hline
		Location & P1 & P2 & P3  \tabularnewline
        \hline  % the coordinate has been rearranged
		Longitude & $120.153173^\circ$ & $120.15488^\circ$  & $120.153894^\circ$   \tabularnewline
        \hline
        Latitude & $30.269884^\circ$ & $30.268726^\circ$  & $30.27096^\circ$    \tabularnewline

        \hline
         P4 & P5 & P6 & P7  \tabularnewline
         \hline
        $120.156252^\circ$ & $120.153905^\circ$  &$120.15936^\circ$  &$120.155162^\circ$   \tabularnewline
         \hline
        $30.270242^\circ$ & $30.27358^\circ$  &$30.273139^\circ$  &$30.278369^\circ$   \tabularnewline

        \hline
        P8 & P9 & P10 & Photo location \tabularnewline
         \hline
        $120.161912^\circ$  &$120.164792^\circ$  & $120.156541^\circ$ & $120.153955^\circ$ \tabularnewline
         \hline
        $30.276465^\circ$  &$30.279437^\circ$  & $30.283932^\circ$ & $30.267191^\circ$ \tabularnewline
        \hline
	\end{tabular}
\end{table}

\begin{table}[!t]\footnotesize
	\caption{Pairwise Sensors Distance (m).}
	\label{Tab:distances}
    \centering \begin{tabular}{|p{0.4cm}|>{\centering}p{0.15cm}>{\centering}p{0.2cm}>{\centering}p{0.3cm}>{\centering}p{0.34cm}>{\centering}p{0.34cm}>{\centering}p{0.34cm}>{\centering}p{0.44cm}>{\centering}p{0.44cm}>{\centering}p{0.44cm}>{\centering}p{0.6cm}|}
		\hline
		& P1 & P2 & P3 & P4 & P5 & P6 & P7 & P8 & P9 & P10 \tabularnewline
        \hline  % the coordinate has been rearranged
		P1& 0 & 113 & 168 & 281 & 464 & 697 & 1012 & 1150 & 1730 & 1625  \tabularnewline
        P2&   &  0  & 211 & 190 & 509 & 603 & 1040 & 1089 & 1537 & 1660  \tabularnewline
        P3&  &     &  0  & 245 & 296 & 575 & 844 & 1000 & 1450 & 1457  \tabularnewline
        P4&  &     &     &  0  & 388 & 413 & 871 & 899 & 1347 & 1510  \tabularnewline
        P5&  &     &     &     &  0  & 464 & 548 & 819 & 1250 & 1161  \tabularnewline
        P6&  &     &     &     &     &  0  & 625 & 486 & 934 & 1170  \tabularnewline
        P7&  &     &     &     &     &     & 0   & 656 & 918 & 613  \tabularnewline
        P8&  &     &     &     &     &     &     &  0  & 448 & 953  \tabularnewline
        P9&  &     &     &     &     &     &     &     &  0  &  877  \tabularnewline
        P10&  &     &     &     &     &     &     &     &     &  0   \tabularnewline
        \hline
	\end{tabular}
\end{table}

\section{Sensor Deployment}
\label{sec:procedure}
This section describes the deployment of our sensors and cameras in order to collect both high-resolution ground-truth data and corresponding images.

\subsection{Sensor Calibration}

Our sensing platform is equipped with humidity and temperature sensors. The precision is 3\% relative humidity (RH) and $\pm$\SI{0.3}{\celsius}, respectively. The PM sensor model is Nova SDS011, which can detect particles with \SI{0.3}{\um} to \SI{10}{\um} diameters according to the scattered light intensity in a specific direction~\cite{grimm2009aerosol}.

\begin{figure}
\centering
\includegraphics[scale=0.41]{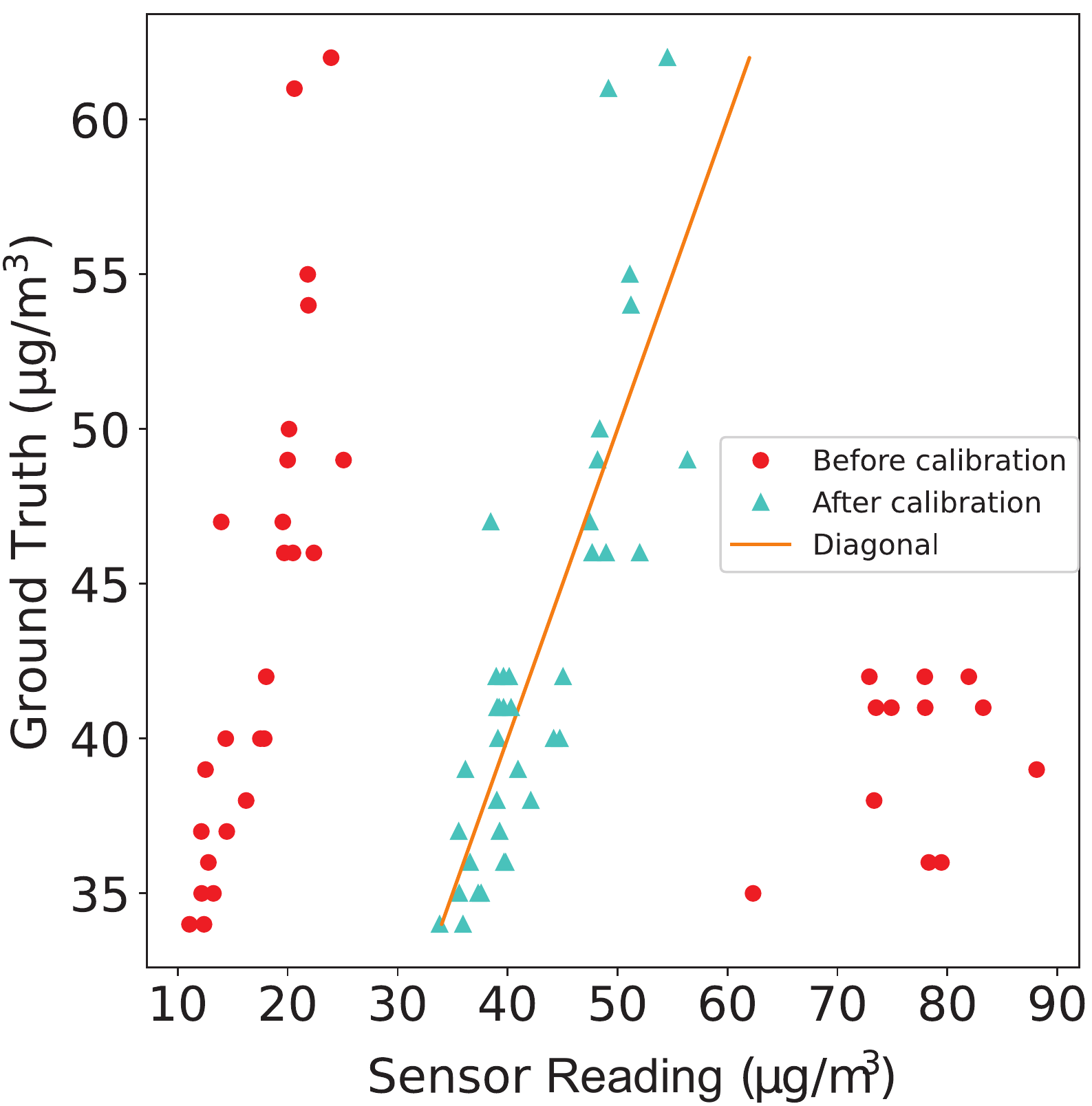}
\caption{Pollution concentration calibration.
}
\label{Fig:calibration}
\end{figure}

We calibrated our particle counting sensors based on the measurements from the air quality monitoring site of Hangzhou Meteorological Bureau in Hemu Primary School using co-location method~\cite{kamionka2006calibration}. Our device is co-located with an air monitoring station, which is equipped with a high-precision sensor. The device collects data for 58 hours continuously. The station data are considered ground truth. We use the least-squares method to fit a quadratic function to the ground truth data. According to the measured concentration, we fit the data into a two-stage piecewise linear function. The fitted function is as follows.
\begin{equation}
	y=\begin{cases}
	1.61x + 16.01 &\text{for } 0 \leqslant x \leqslant 30 \text{ and} \\
	0.13x + 29.48    &\text{for } x > 30, \\
    \end{cases}
\end{equation}
where $x$ is the original value and $y$ is the calibrated one. The calibration data are gathered during  a 2 day period and the result is shown in \autoref{Fig:calibration}. Calibration reduces root mean squared error from \SI{32.74}{\micro\gram\per\cubic\metre} to \SI{3.88}{\micro\gram\per\cubic\metre}, while the variance for all the data over all locations is \SI{9.33}{\micro\gram\per\cubic\metre}.

\subsection{Deployment Details}

\begin{figure} [!t]
\centering
\includegraphics[scale=0.5]{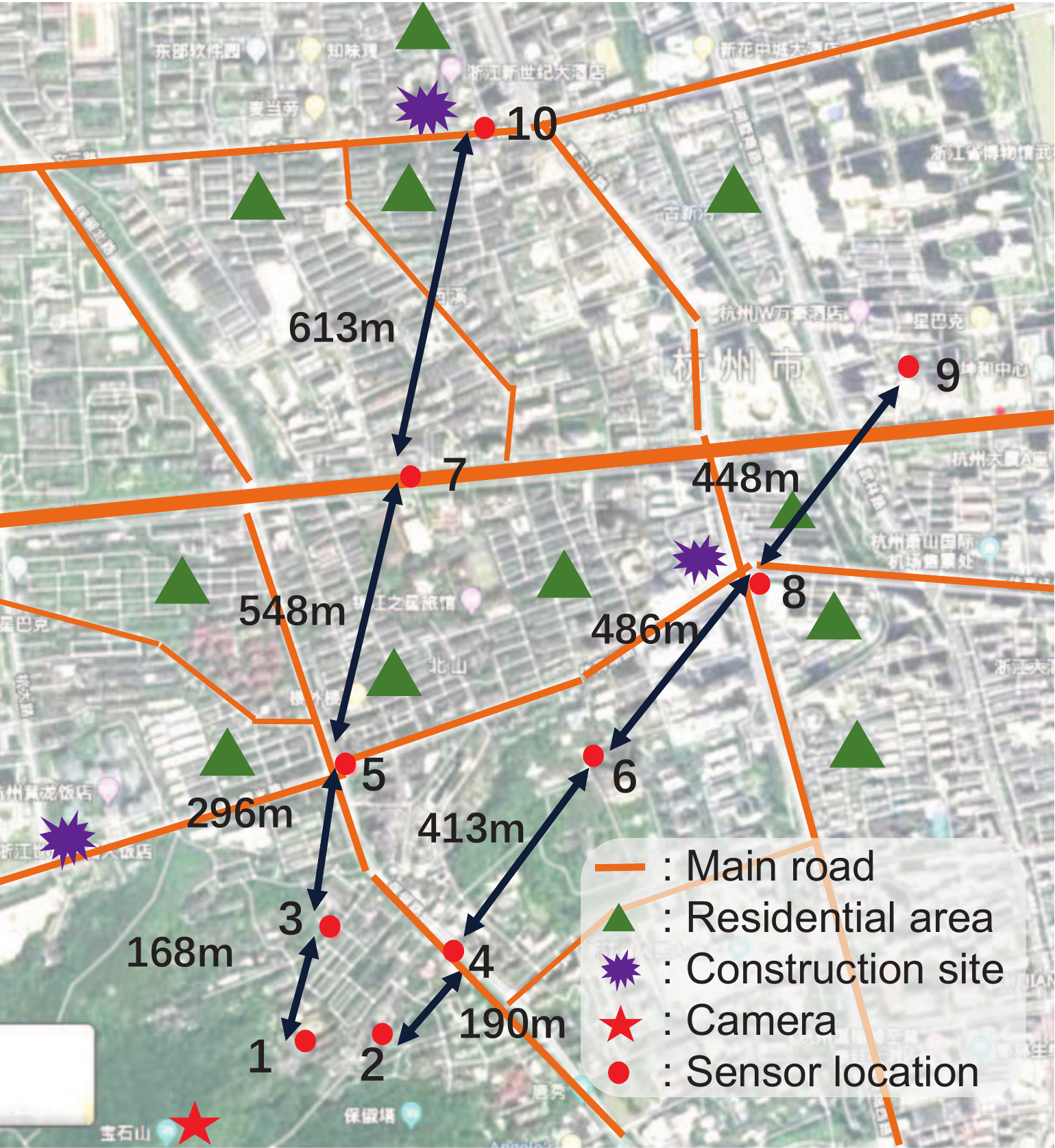}
\caption{The distribution of sensors and pollution sources. Sensor locations are numbered in ascending order according to the distance from the camera.
}
\label{Fig:sensor_locations}
\end{figure}

Our dataset\footnote{Available on \url{https://github.com/implicitDeclaration/HVAQ-dataset/tree/master}.} contains both high-resolution ground truth data and wide-view images. The sensors are placed outdoors and close to living areas. The city center is very crowded and the air quality here affects people more significantly. We deploy our sensors in the urban area of Hangzhou, a city of more than 8 million residents and frequently affected by high PM2.5 concentrations~\cite{liu2015chemical}. Existing research~\cite{jin2017research} shows that in the main urban area of Hangzhou, the sources of PM2.5 are biomass burning/construction dust (41.6\%), vehicle exhaust/metallurgical (metals' production and purification) dust (29.3\%), unknown source (11.2\%), oil combustion (9.8\%), and soil (8.0\%). As shown in \autoref{Fig:sensor_locations}, the main pollution sources are marked on the map and the sensors are located on two straight lines from the observation point. The GPS locations and distances between each pair of sensors are listed in Table~\ref{Tab:GPS} and Table~\ref{Tab:distances}. The sensors are sampled every second and an image is captured every \SI{20}{minutes}. Since the temporal variation of PM concentration is slower than the acquisition rate of the sensors, and the image acquisition rate is also limited by the flight time of quadcopter, images are captured less frequently.

\begin{figure} [!t]
\centering
\includegraphics[scale=0.28]{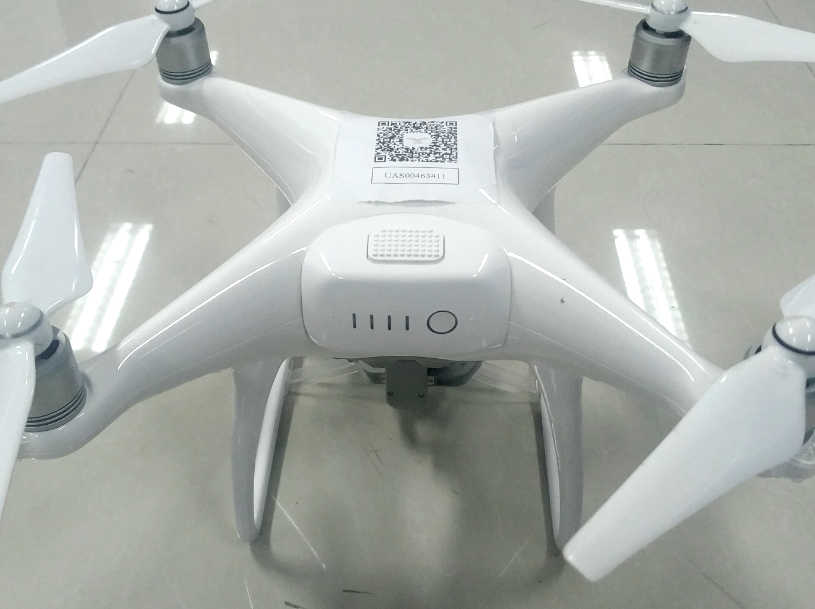}
\caption{The quadcopter (Dji Phantom 4 Pro) used in our deployment.
}
\label{Fig:Dji4}
\end{figure}

To derive a wide-view image covering all sensor locations, we mount a camera on a quadcopter  as shown in \autoref{Fig:Dji4}. We use two approaches to take photos. First, we use a quadcopter equipped with a 4864$\times$3648 camera at \SI{90}{\metre} altitude. Moreover, since the quadcopter has limited carrying and battery capacities, we also take pictures at a fixed location on the mountain top with a resolution of 2592$\times$1936 using an iPad and at a resolution of 4000$\times$3000 using a phone. The parameters of our cameras are listed in \autoref{Tab:camera_info}. The quadcopter camera uses the Sony Exmor R CMOS sensor. The other two cameras types are the Apple iSight and Sony IMX586. 106 images and over 300 thousand PM samples were gathered in three days.

Apparently, the cost of vision-based approach (about \$30) is much less than the total cost of the sensors (about \$670). Thus, predicting air pollution concentrations using images is more convenient and less expensive than particle counters, but requires sophisticated image processing algorithms. \autoref{Tab:equipments} lists our equipment.
\begin{table}[!t]\footnotesize
	\caption{Used equipments.}
	\label{Tab:equipments}
    \centering
	\begin{tabular}{|m{0.8cm}<{\centering}|m{1.5cm}<{\centering}|m{1.5cm}<{\centering}|m{1cm}<{\centering}|m{0.8cm}<{\centering}|m{0.8cm}<{\centering}|}
		\hline
		& Camera & Quadcopter & Sensors & Battery & Platform  \tabularnewline
		\hline
        Price (\$) & 200-460 & 1500 & 3-15 & 12 & 37 \tabularnewline
        \hline
        Model & Quadcopter default camera, iPad, Onplus 7 phone camera & Dji Phantom 4 Pro & PM2.5, PM10, humidity, temperature& 4000 mAh & Raspi 3B+ \tabularnewline
        \hline
	\end{tabular}
\end{table}

\begin{table}[!t]\footnotesize
	\caption{Camera Parameter.}
	\label{Tab:camera_info}
    \centering
	\begin{tabular}{|c|c|c|c|c|}
		\hline
		 Camera & Pixel & Sensor & Aperture   \tabularnewline
        \hline
	     quadcopter & 20,000,000 & Sony Exmor R & f/2.8-f/11  \tabularnewline %$84^\circ$
        \hline
	     pad        & 8,000,000 & iSight  & f/2.4  \tabularnewline
		\hline
         phone      & 48,000,000 & Sony IMX586  & f/1.6  \tabularnewline
		\hline
	\end{tabular}
\end{table}

\section{Dataset Analysis}
\label{sec:dataset}

\textbf{Q1) What is the impact of environmental conditions on measurement accuracy?}

Our data collection period spans three days with varying environmental conditions. The impact of the  environmental conditions determines the corresponding parameters of the calibration functions. Thus, it is important to investigate the relationship between the impact of environmental conditions on measurement accuracy.

\textbf{Q2) What are the spatial variation characteristics of PM2.5 concentration?}

Since the spatial distribution of PM2.5 determines the required sensor density, this information helps us to determine the benefits of increasing the spatial resolution of a pollution sensor network.

\textbf{Q3) How does the correlation of pollution concentrations at two different locations depend on their separation?}

The relationship between PM2.5 concentration and distance provides information relevant to the required number of sensors. For example, if the PM2.5 concentrations at two locations are highly correlated, one sensor might be used to determine the concentrations at both locations.

\textbf{Q4) How much do additional point sensors and vision-based methods improve estimation accuracy?}

This question builds on Q2 and Q3. To evaluate the effectiveness and novelty of our new air quality dataset, it is important to understand how the sensor density and the use of images  relates to the PM2.5 estimation accuracy.

%
%Specifically, we aim to answer the following questions through experimentation and data analysis.
%\begin{enumerate}[]
%
%\item[\textbf{Q1}] What are the impacts of environmental conditions on measurement accuracy?
%
%\item[\textbf{Q2}] What are the spatial variation characteristics of PM2.5 concentration?
%
%\item[\textbf{Q3}] How does pollution concentration correlation change as a function of distance?
%
%\item[\textbf{Q4}] How much do additional point sensors and vision-based techniques improve estimation accuracy?
%
%\end{enumerate}

\subsection{Temperature and Humidity Correlations with Pollution Concentrations}

Environmental factors such as weather conditions can affect sensor readings. For the deployments on Oct. 19 and Nov. 10, we calculate the $R^2$ correlation coefficients for PM2.5 and several environmental factors.

\textbf{A1: Environmental conditions have limited impact.} The results in \autoref{Tab:PM2.5cor} shows that the correlations between PM2.5 and weather factors are insignificant. Wu et al.~\cite{wu2016potential} report that temperature is not significantly correlated to PM concentrations in Hangzhou and PM2.5 concentration is only significantly elevated when RH is higher than 60\%. Moreover, the correlation between environmental factors and PM concentration also depends on the region and season, e.g., Zhu et al.~\cite{zhu2013analysis} report that PM concentration and RH show seasonal correlation. However, since temperature and humidity do not directly affect the measurement process of particle counters, we do not include temperature and humidity in our calibration functions.
%Zhu et al.~\cite{zhu2013analysis} demonstrate that PM2.5 is negatively correlated with relative humidity and shows obvious seasonal differences. The correlation coefficient increased from 0.006 in non-summer to 0.436 in summer. Our data is collected within only one day of July, October and November. Thus seasonal differences is not reflected in our data.

\begin{figure} [!t]
\centering
\includegraphics[scale=0.62]{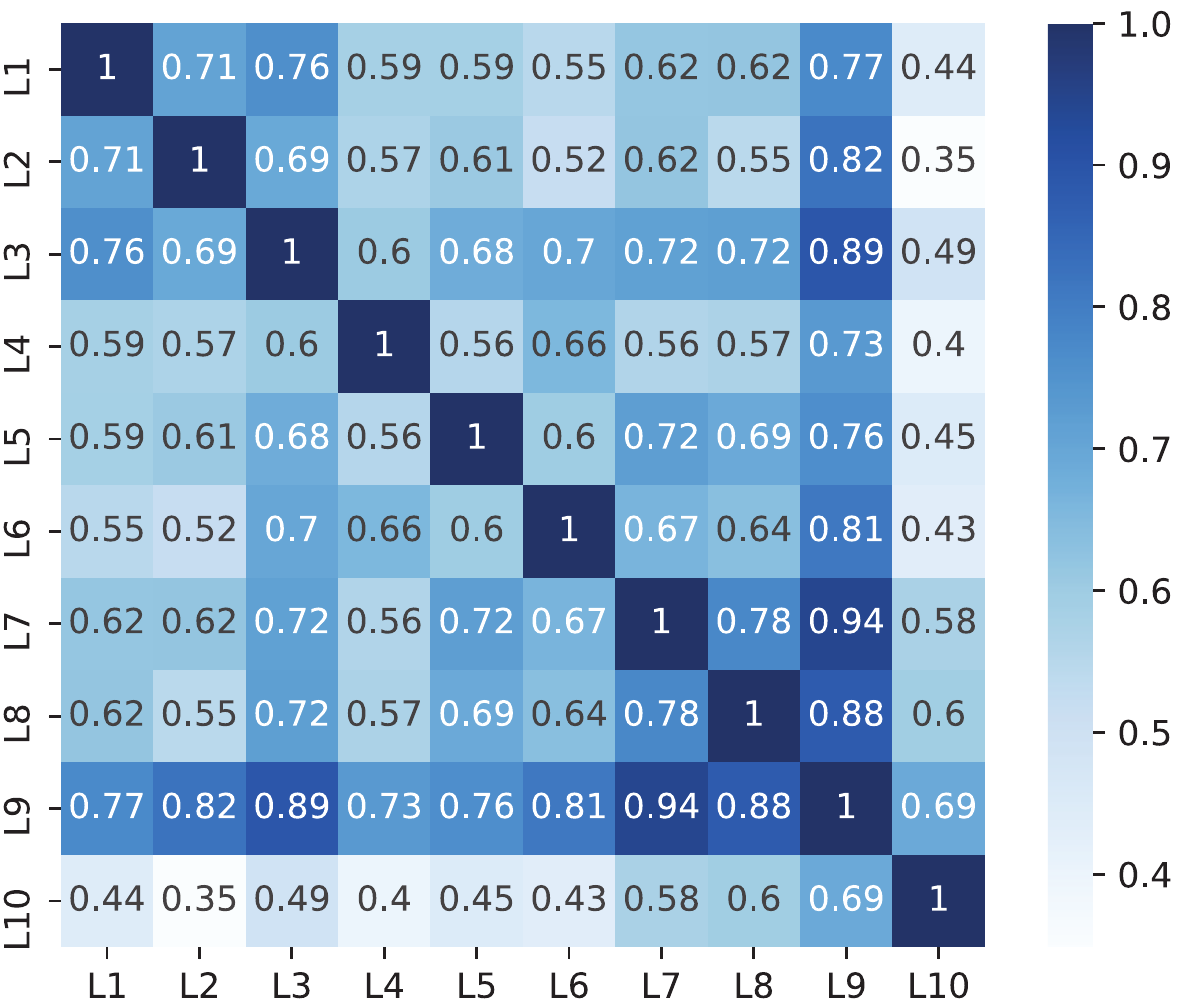}
\caption{The confusion matrix for PM2.5 correlation. The number on the axis represents the number of the corresponding location.}
\label{Fig:correlation_all}
\end{figure}

\begin{table}\footnotesize
\caption{PM2.5 Correlation with PM10 and Environmental Factors on Oct. 19 and Nov. 10.}
\label{Tab:PM2.5cor}
\centering
\begin{tabular}{|c|c|c|c|c|}
\hline
        & PM2.5  & Temperature & Humidity \tabularnewline
\hline
PM2.5   & 1.0   & 0.298 & 0.306 \tabularnewline
\hline
Temperature & 0.298 & 1.0  & 0.808  \tabularnewline
\hline
Humidity    & 0.306 & 0.808 & 1.0  \tabularnewline
\hline
\end{tabular}
\end{table}

\subsection{Correlations of PM Readings}

%(\textmu g/{$\rm m^{3}$})
\begin{table}\footnotesize
\caption{Statistics for PM and Environmental Data.}
\label{Tab:SRA}
\centering
\begin{tabular}{|p{2.5cm}|>{\centering}p{1.4cm}>{\centering}p{1.0cm}>{\centering}p{1.0cm}|>{\centering}p{0.95cm}|>{\centering}p{0.2cm}}
\hline
Data in Different Days  & Standard Deviation &  Data Range  & Average Value & Date \tabularnewline
\hline
PM2.5 (\SI{}{\micro\gram\per\cubic\metre})     & 2.03               & 4.56          & 13.93    & \multirow{1}*{Jul. 24}      \tabularnewline % 07/24 &
%PM10        & 4.48               & 10.04         & 21.99    & ~    \tabularnewline
            \hline
PM2.5 (\SI{}{\micro\gram\per\cubic\metre})      & 2.22               & 5.11          & 25.03    & \multirow{1}*{Jul. 06}      \tabularnewline    % 07/06
%PM10        & 5.45               & 12.66         & 35.80    & ~    \tabularnewline
            \hline
PM2.5 (\SI{}{\micro\gram\per\cubic\metre})      & 6.68               & 22.21         & 56.90    & \multirow{3}*{Oct. 19}      \tabularnewline % 10/19
%PM10        & 17.12              & 52.76         & 79.26    & ~    \tabularnewline
Temperature (\SI{}{\celsius}) & 5.64               & 16.02         & 26.40    & ~    \tabularnewline
RH (\%)    & 12.36              & 37.63         & 45.13    & ~    \tabularnewline
            \hline
PM2.5 (\SI{}{\micro\gram\per\cubic\metre})      & 4.77               & 16.51         & 48.76    & \multirow{3}*{Nov. 10}      \tabularnewline    % 11/10
%PM10        & 19.58              & 67.16         & 82.80    & ~      \tabularnewline
Temperature (\SI{}{\celsius}) & 4.53               & 14.21         & 24.79    & ~      \tabularnewline
RH (\%)   & 10.20              & 34.71         & 40.16    & ~      \tabularnewline
      \hline
\end{tabular}
\end{table}

Our measurements demonstrate that it typically takes more than 10 minutes for concentration to change by \SI{10}{\micro\gram\per\cubic\metre}. Our data sampling period is one second and is considered adequate given the relatively slow concentration change. We quantify the spatial variation of pollution by calculating the standard deviations over all sensors.

\begin{figure} [!t]
\centering
\includegraphics[scale=0.7]{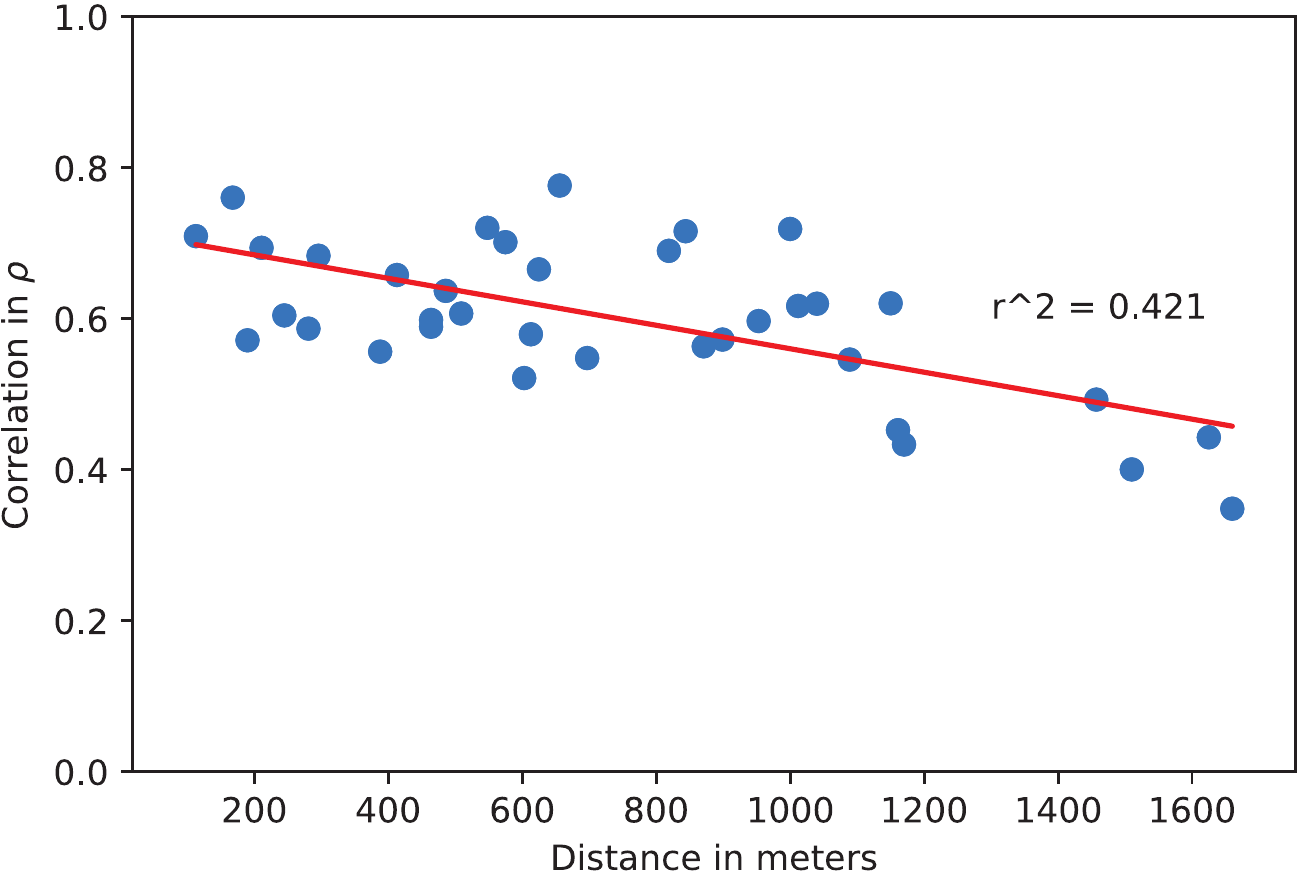}
\caption{Pairwise correlation between sensors for Oct. 19 and Nov. 10 as functions of their distances.}
\label{Fig:cor-dis}
\end{figure}

\textbf{A2: Spatial variation of PM2.5 is high.} As shown in \autoref{Tab:SRA} and \autoref{Fig:cor-dis}, the 2 locations have different concentration and variation trends. Moreover, the large differences indicate that multiple pollution levels coexist in a single image.

\textbf{A3: Pollution concentrations are spatially correlated.} We quantify the correlations for further analysis. \autoref{Fig:cor-dis} shows sensor correlations as a function of pairwise distance. The correlation is measured by the Spearman correlation coefficient:
\begin{equation}
\label{Eq:spearman}
\rho=1-\frac{6\sum d_i^2}{N(N^2-1)},
\end{equation}
where $d_i$ represents the position difference of the paired variables after the two variables are sorted separately and $N$ is the total number of samples. The slope of the fitting line is \SI{0.155}{\per\kilo\metre} and the coefficient of determination ($R^2$) is 0.421, which implies that closer sensors enable more accurate estimates. \autoref{Fig:correlation_all} shows the sensor correlations during the Nov. 10 and Oct. 19 deployments. We expect the correlations to decrease with increasing inter-sensor distances. The deployment results confirm our hypothesis.

\section{Experimental Results}
\label{sec:exp}
In this section, we evaluate the impacts of changing sensor density and using vision-based techniques on estimation accuracy. Specifically, we estimate PM2.5 concentration based on vision-based analysis using transmission information and standard deviation of gray-scale pixel values and compare the performance of several state-of-art estimation algorithms.

\subsection{Experimental Setup}

\begin{figure} [!t]
\centering
\includegraphics[scale=0.3]{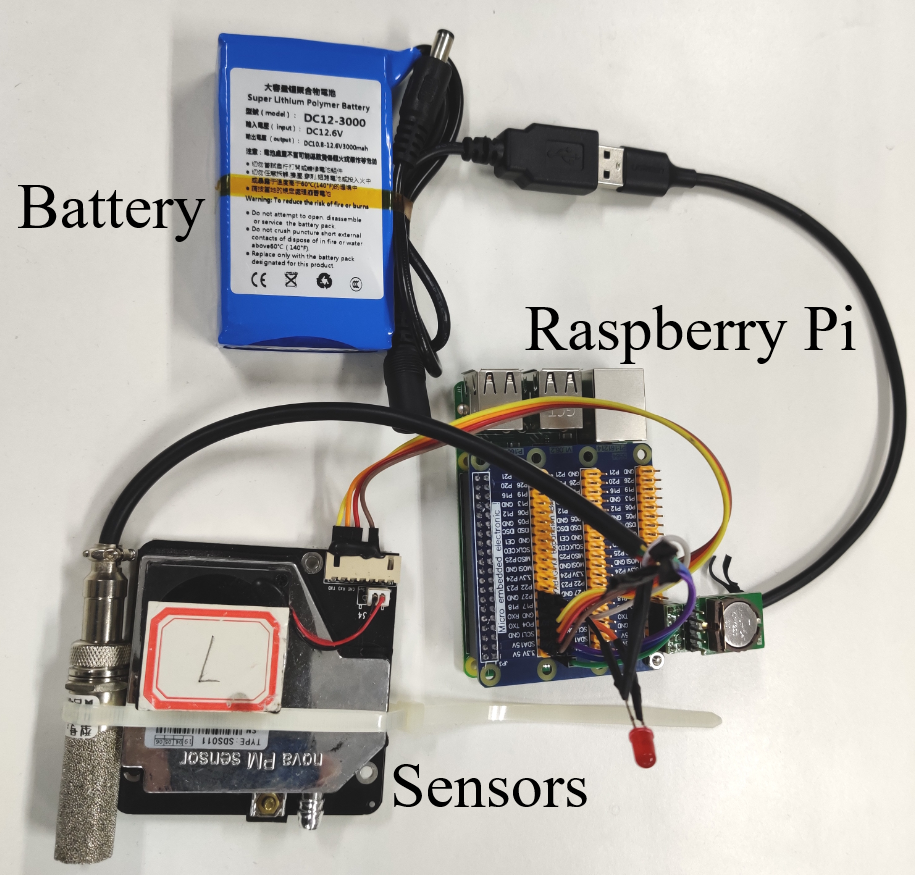}
\caption{The sensing platform consisting of battery, Raspberry Pi, and sensors.
}
\label{Fig:device}
\end{figure}

We design a portable sensing platform to collect, process, and transmit data, as shown in \autoref{Fig:device}. The system battery life is 3.5 hours. The following algorithms are used to estimate pollutant concentration from the measured data.
\begin{enumerate}

\item \textbf{Random forest regression (RFR):} This is a widely used algorithm for regression and classification. It combines multiple weak models to form a strong model with much better performance. Random forests contain multiple, unrelated decision trees. The final output depends on the decisions of all trees.

\item \textbf{Gradient boosting regression (GBR):} This method combines a group of weak learners with low complexity and low training cost. It reduces the problem of overfitting and modifies the weights at each training round to produce a strong learner. Gradient boosting modifies its models based on the gradient descent direction of the loss functions of the previously established models.

\item \textbf{Support vector regression (SVR):} Support vector regression aims to find a regression plane with minimal distance to the dataset. Generally, there is a kernel function mapping data to high-dimensional space for better performance on complex manifolds.

\end{enumerate}
The algorithms are implemented using the Python package \textit{sklearn}. The RFR parameter \textit{n\_estimators} is set to 100 and \textit{criterion} is set to mean squared error. We set the GBR parameter \textit{learning\_rate} to 0.1 and \textit{n\_estimators} to 100, using least squares regression.  We set the SVR parameter \textit{c} to 1.0 and \textit{epsilon} to 0.1, using a radial basis function kernel. The parameters are chosen to maximize the training performance.
We use the following equation to combine sensor readings and image properties:
\begin{equation}
\label{Eq:estimation}
S(x,t)=G(s_1(t),s_2(t),...,s_{10}(t), t_{dcp}(t), \beta_{sd}(t)),
\end{equation}
where $x$ is the location index, $S(x,t)$ is the concentration estimation at time $t$, $s_1,s_2,...,s_{10}$ are the available 10 sensors, $t_{dcp}(t)$ is the transmission information at time $t$, $\beta_{sd}(t)$ is the standard deviation of gray-scale image, and $G$ is the estimation algorithm, which refers to RFR, GBR, or SVR. We use $t_{dcp}(t)$ for low-altitude data and $\beta_{sd}(t)$ for high-altitude data. We later describe how $t_{dcp}(t)$ and $\beta_{sd}(t)$ are obtained for each image in the dataset in section 5.B.

We use data from Jul.\ 24, Oct.\ 19, and Nov.\ 10 for all time stamps and use the mean absolute error (MAE) as the evaluation criteria.
\begin{equation}
\mathit{MAE} = \frac{1}{N}\sum^N_{i=1}|y_i-\hat{y}_i|,
\end{equation}
where $y_i$ is the actual PM2.5 concentration, and $\hat{y}_i$ is the predicted PM2.5 concentration.

There are two classes of images in our dataset. Those in the first class were captured from the ground, on top of a mountain (\SI{78}{\metre}). The second class contains images from a quadcopter flying above the same mountain (\SI{78}{\metre} + \SI{90}{\metre}). We tried to take the images from the quadcopter and mountain at the same angle and keep the images as similar to each other as possible, but there is some (unavoidable) variation in camera orientation (see \autoref{Fig:picture view}).

We divide our data into ``low-altitude'' and ``high-altitude'' subsets according to the image class and evaluate our algorithms separately on the two classes. We analyze 19 images from the high-altitude dataset and 26 images from the low-altitude dataset. Furthermore, each dataset is divided as follows: 75\% of the data and images are randomly selected as a training set and the rest are the testing set. We run the prediction model 50 times per random split of the training and testing datasets.

\subsection{Image Enhanced Concentration Estimation}

\begin{figure} [!t]
\centering
\includegraphics[scale=0.4]{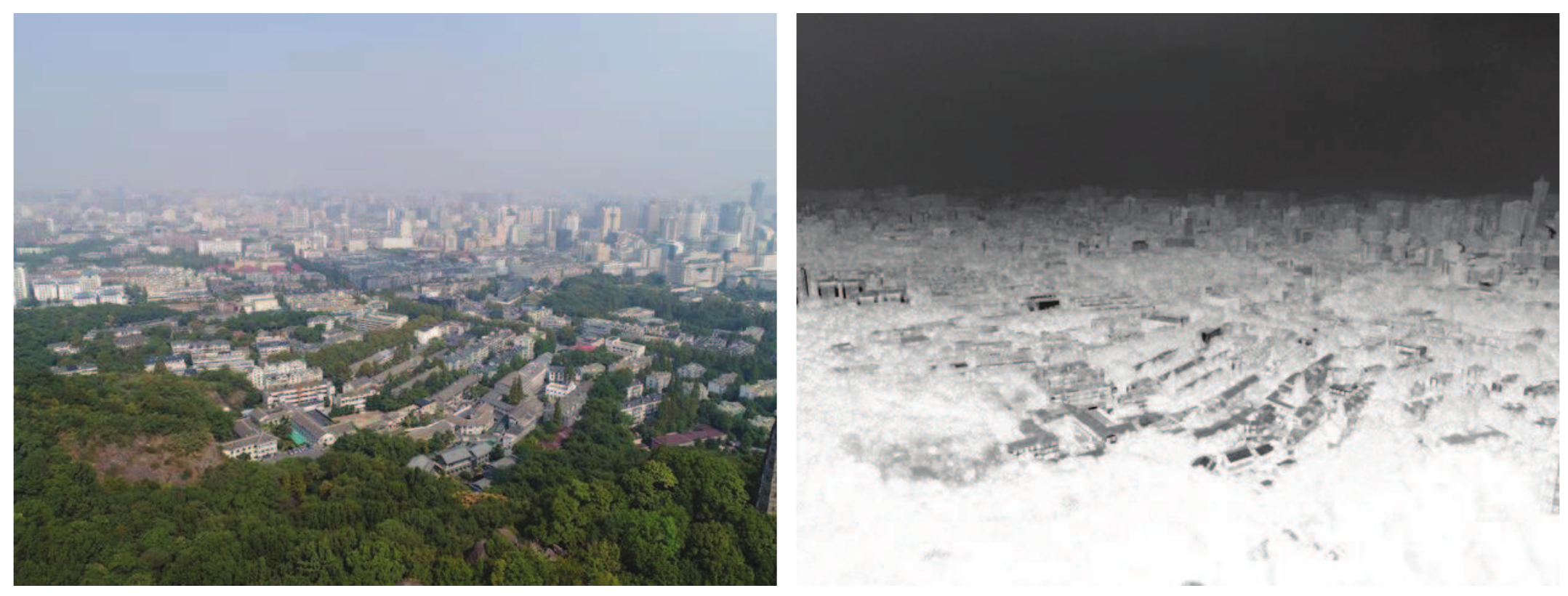}
\caption{Transmission estimation result.
}
\label{Fig:trans_dark}
\end{figure}

Images can be used to estimate PM concentrations in large areas because images can extract the haziness information. We predict PM2.5 concentrations at locations without sensors. Since PM2.5 attenuates light, we estimate PM2.5 concentration in part through the light attenuation coefficient $\beta$ from the haze model. We use the following image features to estimate PM2.5: the dark channel $t_{dcp}(x)$ and the standard deviation $\beta_{sd}$. $t_{dcp}(x)$ is determined using \autoref{Eq:final_trans} and  $\beta_{sd}$ is determined using \autoref{Eq:scattering_final}. \autoref{Fig:trans_dark} is an example of applying the dark channel prior to an image. The left image is the original hazy image, and the right image is the transmission estimate for each pixel using dark channel prior. Note that certain kinds of weather conditions like rain and snow might be misinterpreted as pollution. Our PM2.5 estimation algorithm mainly works in sunny and cloudy conditions, which are common in Hangzhou.

The atmospheric model describing an image influenced by haze follows~\cite{narasimhan2002vision}:
\begin{equation}
\label{Eq:atm_model}
\bm{I}(x) = \bm{J}(x)t(x) + \bm{A}(1-t(x)),
\end{equation}
where $x$ is the pixel location, $\bm{I}$ is the observed image, $\bm{J}$ is the scene radiance (image without any haze), $\bm{A}$ is the atmospheric light, $t$ is the transmission function.

Images with higher air pollution tend to look hazier due to lower transmission and contrast. Hence, image features that correlate to haze level enable pollutant concentration estimation.

\subsubsection{Dark Channel Prior}
The dark channel prior, which has been widely used for haze removal, can be used to estimate the transmission of each image pixel. The dark channel prior method is based on the observation that in most haze-free patches, at least one color channel has some pixels with very low intensities. The dark channel is defined as the minimum of all pixel colors in a local patch and can be calculated using the following equation~\cite{he2010single}:
\begin{equation}
\label{Eq:dark channel}
J_{dark}(x)=\min\limits_{c\in{r,g,b}}\left( \min\limits_{y\in \Omega_{r}(x)}J^c(y) \right),
\end{equation}
where $J^c$ is an RGB channel of $\bm{J}$ and $\Omega_r(x)$ is a local patch centered at $x$ with the size of $15\times 15$. Assume the atmospheric light $\bm{A}$ is given and the transmission in a local patch $\Omega_r(x)$ is constant, taking the minimum operation in the local patch on \autoref{Eq:atm_model}, we have
\begin{equation}
\label{Eq:trans_derivation1}
\min\limits_{y\in \Omega_{r}(x)}\left( I^c(y)\right)=\tilde{t}(x)\min\limits_{y\in \Omega_{r}(x)}\left( J^c(y) \right)+(1-\tilde{t}(x))A^c,
\end{equation}
where $\tilde{t}(x)$ is the patch's transmission. The minimum operation is performed on three color channels independently, it is equivalent to
\begin{equation}
\label{Eq:trans_derivation2}
\min\limits_{y\in \Omega_{r}(x)}\left(\frac{I^c(y)}{A^c}\right)=\tilde{t}(x)\min\limits_{y\in \Omega_{r}(x)}\left(\frac{J^c(y)}{A^c} \right)+(1-\tilde{t}(x)).
\end{equation}

By taking the minimum of three color channels, we have
\begin{equation}
\label{Eq:trans_derivation3}
\begin{split}
\min\limits_{c}\left( \min\limits_{y\in \Omega_{r}(x)}\left(\frac{I^c(y)}{A^c}\right) \right)=&\tilde{t}(x)\min\limits_{c}\left(\min\limits_{y\in \Omega_{r}(x)}\left(\frac{J^c(y)}{A^c} \right) \right) \\
&+(1-\tilde{t}(x)).
\end{split}
\end{equation}

According to the definition of dark channel prior, the dark channel $J_{dark}$ of the haze-free radiance $\bm{J}$ tends to be zero
\begin{equation}
\label{Eq:trans_derivation4}
J_{dark}(x)=\min\limits_{c}\left( \min\limits_{y\in \Omega_{r}(x)}J^c(y)\right)=0.
\end{equation}

Because $A^c$ is always positive, this lead to
\begin{equation}
\label{Eq:trans_derivation5}
\min\limits_{c}\left( \min\limits_{y\in \Omega_{r}(x)}\frac{J^c(y)}{A^c}\right)=0.
\end{equation}

Substituting \autoref{Eq:trans_derivation5} into \autoref{Eq:trans_derivation3}, we can estimate the transmission as follows.
\begin{equation}
\label{Eq:trans_derivation6}
\tilde{t}(x) = 1-\min\limits_{c}\left( \min\limits_{y\in \Omega_{r}(x)}\frac{I^c(y)}{A^c}\right).
\end{equation}

In practice, the atmosphere always contains some haze, which provides depth information. We can optionally keep a small amount of haze by introducing a constant parameter $\omega \ (0<\omega<1)$ into \autoref{Eq:trans_derivation6}
\begin{equation}
\label{Eq:final_trans}
\tilde{t}(x) = 1-\omega\min\limits_{c}\left( \min\limits_{y\in \Omega_{r}(x)}\frac{I^c(y)}{A^c}\right).
\end{equation}

We fix the value of $\omega$ to 0.95 because this approximates the sparse haze present even on relatively clear days. The atmospheric light $A$ is estimated through this procedure: we pick the top 0.1\% brightest pixels in the dark channel and the input image $\bm{I}$ to calculate the atmospheric light. For each hazy image in our dataset, we take the average of the pixel-level transmissions estimated using \autoref{Eq:final_trans}. The resulting average is used for low-altitude data.

\subsubsection{Standard Deviation}
In order to calculate the intensity standard deviation for each image, we convert the RGB image to a gray-scale image, then calculate the standard deviation of all the pixel intensities. The standard deviation is closely related to haze density~\cite{park2013fast}. The scattering coefficient is a measure of haze density. The higher the scattering coefficient, the higher the haze density. The transmission function is
\begin{equation}
\label{Eq:transmission}
t(x)=e^{\beta d(x)},
\end{equation}
where $\beta$ is the scattering coefficient and $d$ is the depth. Substituting \autoref{Eq:transmission} into \autoref{Eq:atm_model}, we have
\begin{equation}
\label{Eq:scattering1}
\bm{I}_g(x) = \bm{J}(x)e^{\beta d(x)}+ \bm{A}(1-e^{\beta d(x)}).
\end{equation}

The variance of a gray-scale image is
\begin{equation}
\begin{split}
\label{Eq:scattering2}
\sigma_{I_g}^2 &= \frac{1}{N}\sum_{i=1}^{N}(I_g(i)-\frac{1}{N}\sum_{j=1}^{N}I_g(j))^2 \\
&=e^{2\beta d(x)}\frac{1}{N}\sum_{i=1}^{N}(J(i) - \frac{1}{N}\sum_{j=1}^{N}J(j))^2,
\end{split}
\end{equation}
where $I_g$ is the gray-scale image and $N$ is the number of pixels in the image. When $\beta=0$, we have
\begin{equation}
\label{Eq:scattering3}
\sigma_0^2 = \frac{1}{N}\sum_{i=1}^{N}(J(i)-\frac{1}{N}\sum_{j=1}^{N}J(j))^2.
\end{equation}
Combining \autoref{Eq:scattering3} and \autoref{Eq:scattering2} yields
\begin{equation}
\label{Eq:scattering4}
\sigma_{I_g} = e^{-2\beta} \sigma_0^2.
\end{equation}
After taking the logarithm of both sides, the scattering coefficient can be expressed as
\begin{equation}
\label{Eq:scattering5}
\beta=\rm{ln} \ \sigma_0-\rm{ln} \ \sigma_{I_g}.
\end{equation}
Since $\sigma$ is changed at each image, \autoref{Eq:scattering5} can be expressed using the first-order Taylor Series approximation
\begin{equation}
\label{Eq:scattering6}
\beta= 1 + \rm{ln} \ \sigma_0 - \sigma_{I_g}.
\end{equation}
When $\beta=0$, the variance of the scene radiance approximates 1. Thus we have
\begin{equation}
\label{Eq:scattering_final}
\beta=1-\rm{ln} \ \sigma_{I_g}.
\end{equation}

Therefore we can estimate the concentration using the standard deviation of the gray-scale image. The standard deviation is used for high-altitude data.

\subsection{Concentration Estimation Results}
\begin{figure} [!t]
\centering
\includegraphics[scale=0.5]{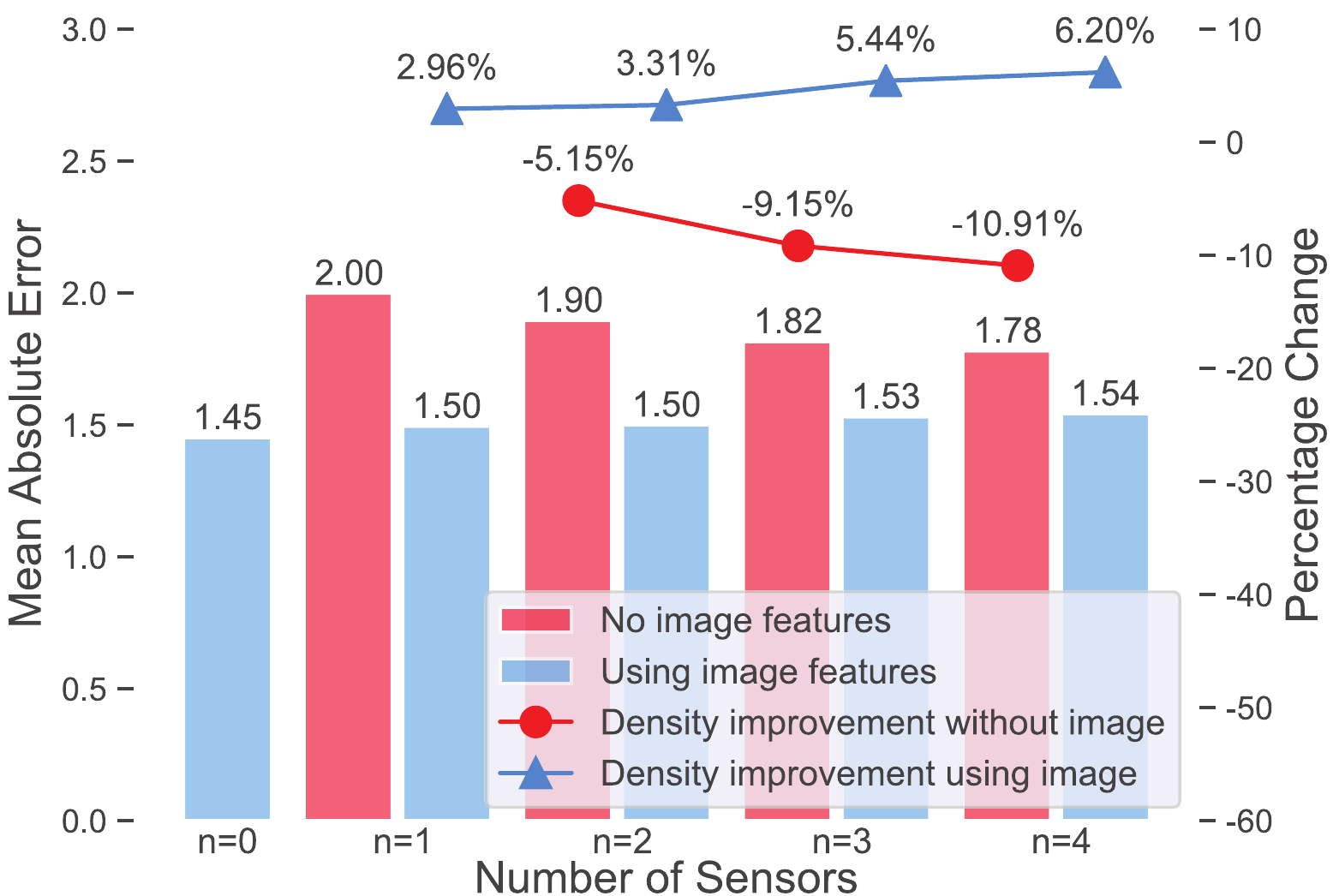}
\caption{The relationship between mean average error and sensor density for Gradient Boosting Regression on high-altitude data.}
\label{Fig:MAE_drone}
\end{figure}

\begin{figure} [!t]
\centering
\includegraphics[scale=0.5]{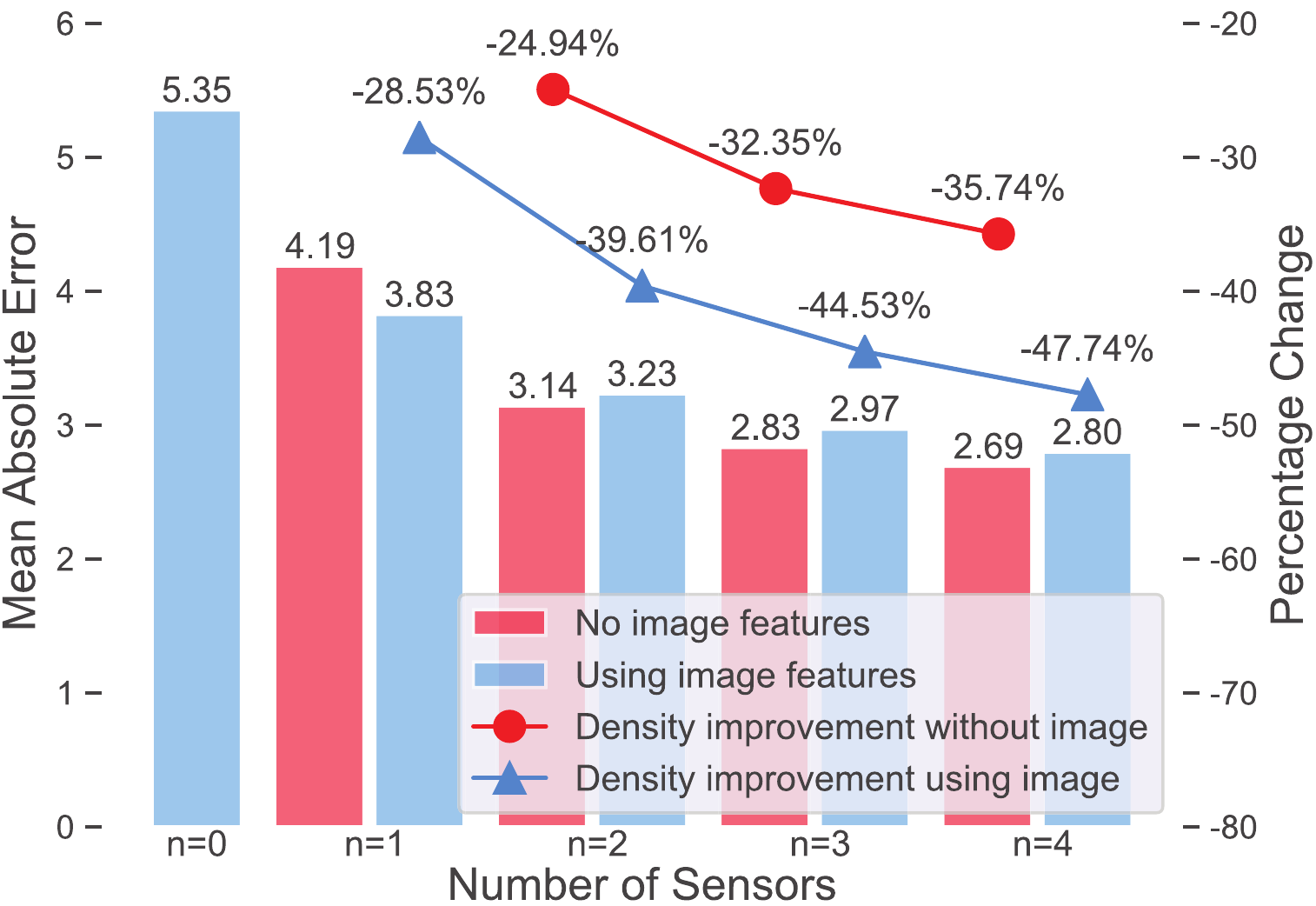}
\caption{The relationship between mean average error and sensor density for Gradient Boosting Regression on low-altitude data.}
\label{Fig:MAE_phone}
\end{figure}

\begin{figure} [!t]
\centering
\includegraphics[scale=0.45]{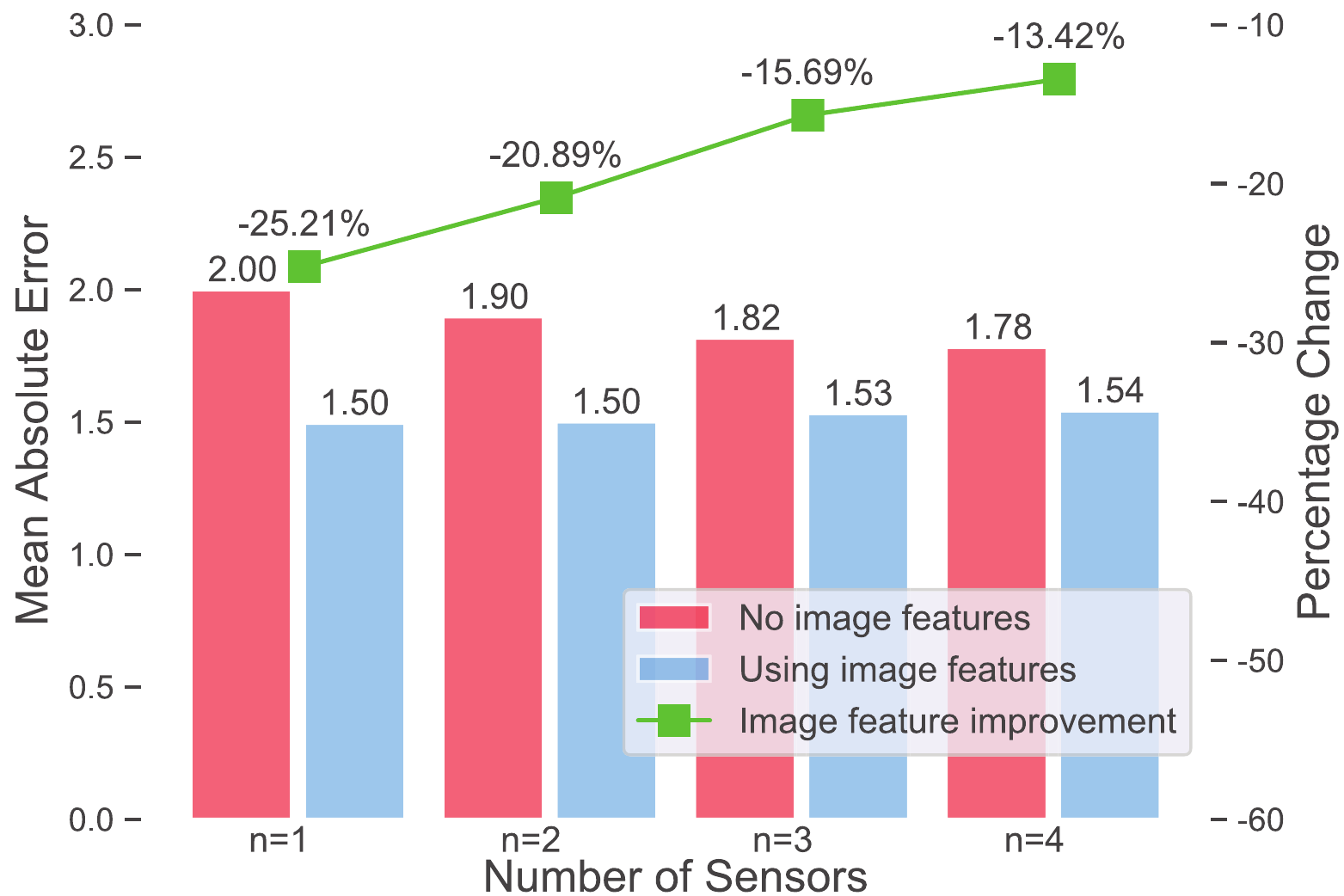}
\caption{The relationship between mean average error and using images for Gradient Boosting Regression on high-altitude data.}
\label{Fig:MAE_Image_drone}
\end{figure}

\begin{figure} [!t]
\centering
\includegraphics[scale=0.45]{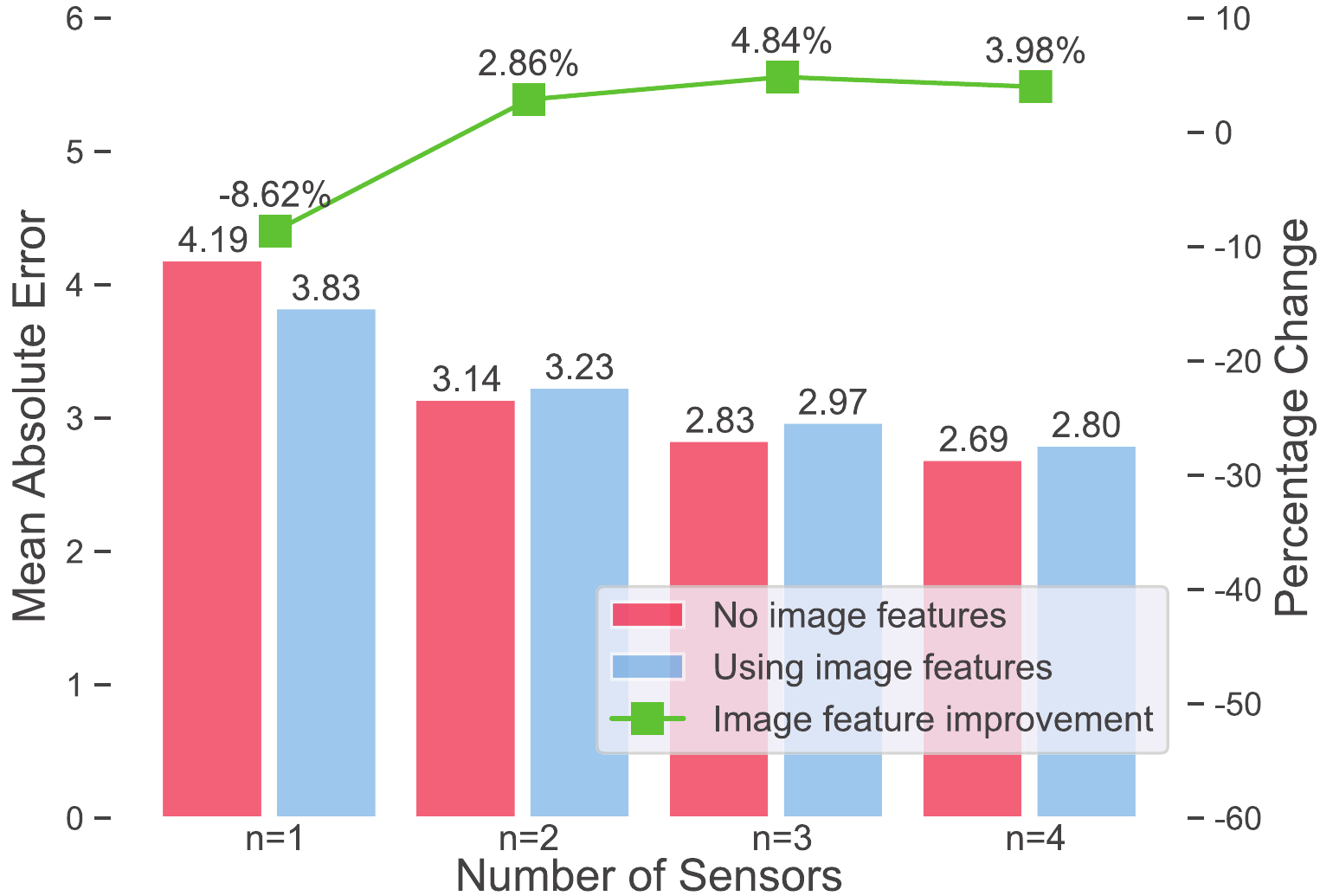}
\caption{The relationship between mean average error and using images for Gradient Boosting Regression on low-altitude data.}
\label{Fig:MAE_Image_phone}
\end{figure}

For each available sensor deployment location, we speculatively remove one or more sensor's data and use estimation techniques with access to the remaining sensors to infer concentration(s), thereby allowing comparison with ground truth measurements. We investigate the impact of the number of sensors on the estimation accuracy by using all possible combinations of speculatively removed sensors and averaging the results. We also consider the impact of using image data on estimation accuracy.

We plotted the results from the best-performing algorithm: GBR. The density improvements in Figures~\ref{Fig:MAE_drone} and \ref{Fig:MAE_phone} are compared with the $n=0$ (no images) case. If image features are used, the improvements are compared with the $n=1$ case. Note that the lower MAE is, the better the estimation accuracy. Thus a negative change indicates improvement. The bars in Figures~\ref{Fig:MAE_drone}--\ref{Fig:MAE_Image_phone} represent the MAE resulting from using different numbers of sensors. The lower bars indicate higher accuracies. The lines in Figures~\ref{Fig:MAE_drone}--\ref{Fig:MAE_Image_phone} indicate percentage change to MAE in different cases (e.g., using images or more sensors). The lower the percentage change, the larger the improvement, and positive percentage changes imply (undesirable) increases in MAE.

As sensor density increases, the estimation MAE decreases. This implies that accuracy improves with sensor density, even at densities much higher than those of modern stationary sensor deployments. Moreover, as shown in \autoref{Fig:cor-dis}, sensor correlations decrease with increasing distance. This is the reason estimation accuracy improves with increasing sensor density. Using the four nearest sensors instead of one sensor improves estimation accuracy by 23.3\% on average without using images, and 20.75\% when using images. The fact that increasing sensor density improves accuracy less when images are available does not imply that images are unhelpful. In contrast, it implies that using images allows higher accuracy when few sensors are available, leaving less potential for improvement if sensor density is later increased.

\textbf{A4 Vision-based techniques significantly improve estimation accuracy.} As shown in Figures~\ref{Fig:MAE_Image_drone} and \ref{Fig:MAE_Image_phone}, PM2.5 concentration prediction accuracy improves when images are used. In the case of $n=0$, we average all the available concentrations. The MAE is \SI{20.821}{\micro\gram\per\cubic\metre} for high-altitude data and \SI{6.929}{\micro\gram\per\cubic\metre} for low-altitude data. The high-altitude images enable higher accuracy because some sensor locations are occluded in the low-altitude images. When images are used, MAE drops to \SI{1.45}{\micro\gram\per\cubic\metre} and \SI{5.35}{\micro\gram\per\cubic\metre} respectively. For the case where $n=1$, when we use the PM2.5 concentrations of the nearest available sensor for estimation image data improves prediction accuracy by 16.9\% on average. The benefits of using images are greatest when the fewest particle counters are used.

To determine whether the improvement is systemic and statistically significant, we use the Kolmogorov-Smirnov test. We compare the data for the MAE without using images that using images for each number of sensors ($n = 1, ..., 4$) to determine whether the two data sets have the same distribution. The test is non-parametric and requires no knowledge about the distribution of data.

We determine that there is a statistically significant difference between the distribution of the MAE when using images and the distribution without using images, i.e., the difference between the two distributions is not due only to chance or noise but due to a genuine improvement in accuracy when using the image data. If the p-value is below 0.10, we can reject the null hypothesis and conclude that using images does improve our results. As shown in \autoref{Tab:pvalue1} and \autoref{Tab:pvalue2}, on low-altitude data, the p-values of GBR in all cases are less than 0.10, and on high-altitude data, GBR's p-values are less than 0.10. As a result, we are confident that using images improves the estimates.

\begin{table}\footnotesize
\caption{P-Values of GBR on High-Altitude Data.}
\label{Tab:pvalue1}
\centering
\begin{tabular}{|>{\centering}p{1.8cm}|>{\centering}p{1.8cm}|>{\centering}p{1.8cm}|>{\centering}p{1.4cm}|>{\centering}p{1.2cm}|}
\hline
        \multicolumn{4}{|c|}{GBR} \tabularnewline
\hline
        n=1 & n=2 & n=3 & n=4 \tabularnewline
\hline
        $ <0.001$ & $ <0.001 $ & $<0.001$ & $0.064$ \tabularnewline
\hline
%        \multicolumn{4}{|c|}{SVR} \tabularnewline
%\hline
%         n=1 & n=2 & n=3 & n=4 \tabularnewline
%  \hline
%         $<0.001$ & $<0.001$ & $<0.001$ & $0.010$ \tabularnewline
%\hline
%        \multicolumn{4}{|c|}{RFR} \tabularnewline
%\hline
%         n=1 & n=2 & n=3 & n=4 \tabularnewline
%\hline
%         $0.036$ & $0.041$ & $0.823$ & $0.930$ \tabularnewline
%\hline
\end{tabular}
\end{table}

\begin{table}\footnotesize
\caption{P-Values of GBR on Low-Altitude Data.}
\label{Tab:pvalue2}
\centering
\begin{tabular}{|>{\centering}p{1.6cm}|>{\centering}p{1.6cm}|>{\centering}p{1.8cm}|>{\centering}p{1.8cm}|>{\centering}p{1.8cm}|}
\hline
        \multicolumn{4}{|c|}{GBR} \tabularnewline
\hline
        n=1 & n=2 & n=3 & n=4 \tabularnewline
\hline
        $0.068$ & $<0.001$ & $<0.001$ & $<0.001$ \tabularnewline
\hline
%        \multicolumn{4}{|c|}{SVR} \tabularnewline
%\hline
%         n=1 & n=2 & n=3 & n=4 \tabularnewline
%\hline
%         $0.106$ & $0.012$ & $<0.001$ & $<0.001$ \tabularnewline
%\hline
%        \multicolumn{4}{|c|}{RFR} \tabularnewline
%\hline
%         n=1 & n=2 & n=3 & n=4 \tabularnewline
%\hline
%         $0.009$ & $0.069$ & $0.001$ & $0.002$ \tabularnewline
%\hline
\end{tabular}
\end{table}

Certain fixed-location images show slightly negative results for the GBR method. Since the fixed-location images are taken from a low altitude, some of the sensor locations are blocked by buildings. For the quadcopter images taken at a higher altitude, all estimation techniques improve accuracy. In general, images decrease MAE by 8.44\% on average, when $n\leqslant 1$, adding a camera to collect images helps more than adding more sensors.

Increasing sensor density and using images change the relative accuracy of the estimation algorithms. When we use only one sensor and no images, RFR has lower MAE than GBR on low-altitude data. When the sensor density is increased ($n=4$) and images are used, GBR outperforms RFR. This result demonstrates that it is important to evaluate estimation algorithms using appropriate sensor densities and access to image data. A sparsely deployed and less accurate sensor network can lead to false conclusions about pollution concentrations, and about which pollution concentration estimation algorithms are most accurate.

% area 2.45km^2
To summarize, higher sensor densities and image data both improve estimation accuracy, and adding image data has a similar effect to increasing particle counter density by 0.61 sensors \SI{}{\per\square \kilo \metre}. Of the three estimation techniques evaluated, GBR had the highest accuracy with MAE=\SI{1.45}{\micro\gram\per\cubic\metre}. In particular, our developed method is unlikely to work as well on night-time images and images with adverse weather conditions such as rain and snow. The main limitation of our dataset is that it does not contain nighttime images.

\section{Conclusion}
\label{sec:conclusion}

This paper presents a PM dataset with high spatial and temporal resolution. In contrast with existing datasets, HVAQ contains images covering the locations of stationary point sensors, making it suitable for evaluating and validating vision-based pollution estimation algorithms. Through our analysis, we find that (1) the estimation accuracy can be improved significantly using vision-based techniques; (2) the spatial pollutant distribution is spatially correlated; (3) spatial variation of PM2.5 is high; and (4) temperature and humidity had limited impact on PM concentration in our dataset. We also evaluate our data using state-of-art prediction methods. Accuracy correlates with density with a coefficient of \SI{0.2875}{\micro\gram\per\cubic\metre} MAE per sensor and vision-based estimation improves accuracy by \SI{0.1813}{\micro\gram\per\cubic\metre} MAE, on average.

% or


\section*{Acknowledgment}
This work was supported in part by NSF under award CNS-2008151 and in part by the National Natural Science Foundation of China (61572439) and Zhejiang Provincial Natural Science Foundation of China (LY18F030021, LR19F030001).
The authors would like to thank Zhuangzhuang Liu, Qing Yuan, Qiang Zheng, Wencheng Liu, Chengchao Zhu, Lijie Sun, Xinhui Zhang, Xujie Song, and Jiakang Zhao, who helped with the sensor deployments.

\ifCLASSOPTIONcaptionsoff
  \newpage
\fi

%\balance

\bibliographystyle{IEEEtranTIE}
\bibliography{BIB_17-TIE-2179}\ %IEEEabrv instead of IEEEfull


%\vfill


\end{document}